\documentclass{article}
\usepackage{amssymb}

\usepackage[final]{corl_2025} 

\usepackage{graphicx}
\usepackage{booktabs}
\usepackage{amsmath}
\usepackage{amsfonts}
\usepackage{makecell}
\usepackage{tabularx}
\usepackage{vcell}
\usepackage{wrapfig,lipsum,booktabs}
\usepackage{multirow}
\usepackage{xspace}
\usepackage{caption}
\usepackage{subcaption}
\usepackage{wrapfig}

\usepackage{booktabs}      
\usepackage{graphicx}      
\usepackage{hyperref}  
\usepackage{xspace}
\usepackage{siunitx}

\usepackage[table,xcdraw,dvipsnames]{xcolor}

\usepackage[para]{threeparttable}
\usepackage{tabularx}
\usepackage{booktabs}
\usepackage{dsfont}
\usepackage{bm}
\usepackage{float}

\definecolor{lightgray}{gray}{0.9}

\newcommand{\method}{HuB\xspace}

\newcolumntype{C}{>{\centering\arraybackslash}X} 
\newcolumntype{L}{>{\raggedright\arraybackslash}X}

\newlength\savewidth

\renewcommand{\paragraph}[1]{\vspace{1.25mm}\noindent\textbf{#1}}

\newcolumntype{x}[1]{>{\centering\arraybackslash}p{#1pt}}
\newcolumntype{y}[1]{>{\raggedright\arraybackslash}p{#1pt}}
\newcolumntype{z}[1]{>{\raggedleft\arraybackslash}p{#1pt}}

\usepackage{tikz}

\usepackage[capitalize]{cleveref}
\crefname{section}{Sec.}{Secs.}
\Crefname{section}{Section}{Sections}
\Crefname{table}{Table}{Tables}

\title{\method: Learning Extreme Humanoid Balance}

\author{
Tong Zhang$^{1,2}$\thanks{Equal contribution} \quad 
Boyuan Zheng$^{1}$\footnotemark[1] \quad 
Ruiqian Nai$^{1,2}$ \quad 
Yingdong Hu$^{1,2}$ \quad 
Yen-Jen Wang$^{3}$ \\
\textbf{Geng Chen}$^{4}$ \quad 
\textbf{Fanqi Lin}$^{1,2}$ \quad 
\textbf{Jiongye Li}$^{1}$ \quad 
\textbf{Chuye Hong}$^{1}$ \quad 
\textbf{Koushil Sreenath}$^{3}$ \quad 
\textbf{Yang Gao}$^{1,2}$\thanks{Corresponding author}
\vspace{1mm}
\\
$^1$Tsinghua University\quad  $^2$Shanghai Qi Zhi Institute\quad  $^3$UC Berkeley \quad $^4$UC San Diego \\
}

\begin{document}
\maketitle


\begin{figure}[ht]
    \centering
    \vspace{-1.1cm}
    \includegraphics[width=.87\textwidth]{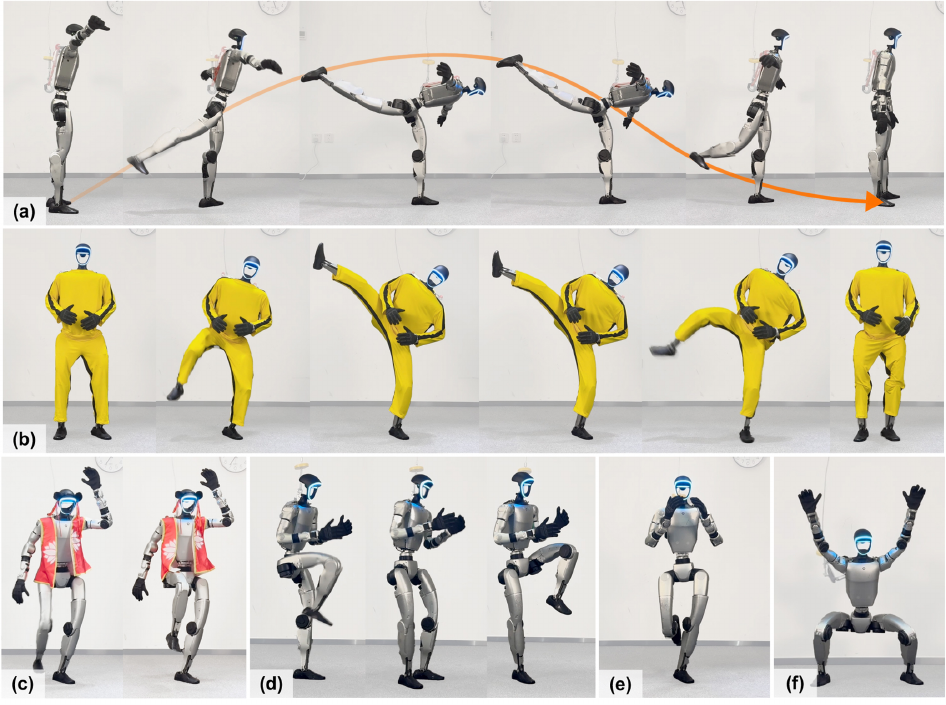}
    \caption{\small{\textbf{Extreme Balance Tasks.} \method enables humanoids to perform extreme quasi-static balance tasks with high stability. (a) \texttt{Swallow Balance}: holding a challenging T-shaped pose with the torso extended horizontally; (b) \texttt{Bruce Lee’s Kick}: executing a high kick with full leg extension while balancing on one foot; (c) \texttt{Ne Zha Pose}: a martial arts-inspired one-legged stance with a raised arm; (d) \texttt{High Knees}; (e) \texttt{Single-Leg Stand}; (f) \texttt{Deep Squat}. Videos are available at: \href{hub-robot.github.io}{hub-robot.github.io}}}
    \vspace{-0.2cm}
    \label{fig:teaser}
\end{figure}

\vspace{-0.3cm} 
\begin{abstract}
The human body demonstrates exceptional motor capabilities—such as standing steadily on one foot or performing a high kick with the leg raised over 1.5 meters—both requiring precise balance control. While recent research on humanoid control has leveraged reinforcement learning to track human motions for skill acquisition, applying this paradigm to balance-intensive tasks remains challenging. In this work, we identify three key obstacles: instability from reference motion errors, learning difficulties due to morphological mismatch, and the sim-to-real gap caused by sensor noise and unmodeled dynamics. To address these challenges, we propose \textbf{\method} (\textbf{Hu}manoid \textbf{B}alance), a unified framework that integrates \textit{reference motion refinement}, \textit{balance-aware policy learning}, and \textit{sim-to-real robustness training}, with each component targeting a specific challenge. We validate our approach on the Unitree G1 humanoid robot across challenging quasi-static balance tasks, including extreme single-legged poses such as \texttt{Swallow Balance} and \texttt{Bruce Lee's Kick}. Our policy remains stable even under strong physical disturbances—such as a forceful soccer strike—while baseline methods consistently fail to complete these tasks. Project website: \href{hub-robot.github.io}{hub-robot.github.io}. 
\end{abstract}



\section{Introduction}
\label{sec:intro}	

Developing humanoid robots that can emulate the versatility, agility, and robustness of human movement in complex, unstructured environments has long been a central pursuit in robotics research~\cite{hirai1998development, johnson2015team,kuindersma2016optimization, he2024learning, fu2024humanplus, he2025asap}. Achieving this vision requires not only the ability to execute diverse motor skills, but also the capacity to maintain balance under challenging conditions. Studies in neuroscience and motor control suggest that human balance relies on intricate sensorimotor loops involving the vestibular system, proprioception, and high-level planning~\cite{fitzpatrick2004probing, peterka2002sensorimotor}, making it a particularly demanding aspect of motor control to replicate in robotics. This difficulty is exemplified by the \texttt{Swallow Balance} task shown in \Cref{fig:teaser}, in which a humanoid must maintain stability in an extreme single-legged pose with the upper body extended horizontally. Such movements require full-body coordination, precise control of the center of mass, and robustness to perturbations—highlighting the demanding nature of humanoid balance.

In recent work on learning-based humanoid control~\cite{he2024learning, fu2024humanplus, he2024omnih2o, he2024hover, cheng2024expressive, ji2024exbody2,he2025asap}, a common approach for enabling humanoids to perform diverse motions is to train a control policy to track reference poses.
The standard pipeline typically begins by obtaining human poses either from video-based motion capture algorithms~\cite{shin2024wham, wang2024tram} or marker-based motion capture systems. These poses are then retargeted to humanoid-specific reference motions. Next, a control policy is trained in simulation to track these reference motions, and finally, the trained policy is deployed to real-world hardware.
However, this pipeline faces significant challenges when applied to complex balancing tasks. In the following, we identify these challenges and present our proposed solutions to address them.

\textbf{Challenges due to Reference Motion Errors.} For tracking-based methods, the successful execution of high-precision balancing critically depends on the accuracy of the reference motion. However, video-based motion capture algorithms~\cite{shin2024wham} often introduce significant errors, and although marker-based motion capture systems offer better precision, they are impractical for Internet videos or consumer-grade recordings. Moreover, optimization-based retargeting~\cite{he2024learning} can further degrade reference quality due to non-convex optimization, imperfect model alignment, and a lack of temporal continuity constraints. These issues can lead to artifacts such as foot sliding even during stationary phases, which cannot be tolerated in demanding balance tasks. These inaccuracies pose substantial challenges for humanoids attempting to perform complex balancing tasks. To address this, we develop a pipeline that leverages carefully designed initialization to accelerate retargeting convergence and incorporates post-processing techniques to enhance physical plausibility and transition stability.

\textbf{Challenges in Training Balance Policies.} Even with relatively accurate reference data, training balance policies still presents significant challenges. Due to morphological differences between the human body and the humanoid, their centers of mass (COM) do not necessarily align. As a result, strictly tracking the reference motion does not always lead to stable equilibrium for the humanoid. To address this, we relax the policy’s tracking objective, allowing it to explore more stable behaviors near the reference trajectory. In addition, to regulate the policy’s motion and encourage physically plausible behavior, we introduce a set of shaping rewards. These design choices enable the policy to discover balance strategies better suited to the humanoid’s own dynamics.

\textbf{Challenges in Sim-to-Real Transfer.} The sim-to-real gap is a fundamental challenge in simulation-based robot learning, and becomes particularly problematic in complex balance tasks. 
In the real world, sensors—especially IMUs and visual-inertial odometry (VIO) systems—are often noisy, which leads to inaccurate policy inputs. This subsequently causes jitter in the action outputs and can trigger a vicious feedback loop of instability.
Moreover, our experiments show that prior tracking-based methods~\cite{he2024learning,he2024omnih2o} often cause humanoids to wobble or jitter during real-world balance tasks. These phenomena primarily arise from modeling discrepancies between simulation and reality, particularly in the simulation of ground contact and frictional interactions. To improve robustness under the sim-to-real gap, we adopt localized reference tracking to eliminate VIO dependence, introduce IMU-centric observation perturbation to model realistic sensor noise, and apply high-frequency external pushes to approximate real-world jitter effects, thereby bridging the sim-to-real gap and enhancing deployment robustness.

The strategies outlined above constitute a comprehensive framework for addressing the inherently complex challenge of balance maintenance in humanoid robots. We validated \method on the Unitree G1 humanoid robot, and the experiments demonstrate that our method enables the robot to perform highly challenging balance tasks, such as \texttt{Swallow Balance} and \texttt{Bruce Lee's Kick}, as illustrated in \Cref{fig:teaser}. In contrast, tracking-based baselines consistently fail to accomplish these tasks, either losing balance and falling, or abandoning single-leg motions. Ablation studies further validate the necessity of each component of our approach. Furthermore, \method exhibits rapid adaptation and strong robustness against external disturbances, such as a forceful strike from a soccer ball, and enables humanoids to successfully complete 10 consecutive executions within a single continuous rollout without any intervention or resets.

\section{Related Work}
\label{sec:related_work}

\textbf{Humanoid Balance Control.}
Maintaining balance is a fundamental capability for humanoids. Classical approaches typically adopt model-based control, including feedback-based~\cite{kajita2003biped, pratt2006capture, stephens2007integral, koolen2012capturability} and optimization-based methods~\cite{atkeson2007multiple, stephens2010dynamic, ott2011posture}. While effective in structured settings, they often rely on accurate dynamics modeling and struggle under uncertainty. More recently, learning-based approaches have leveraged reinforcement learning (RL) for dynamic stepping~\cite{peng2017deeploco, he2025asap}, push recovery~\cite{duburcq2022reactive}, standing-up motions~\cite{he2025learning, huang2025learning, zhuang2025embrace}, and balancing with uncertain contacts~\cite{yang2018learning}. However, prior works primarily focus on locomotion or transient stabilization rather than sustained quasi-static balance under extreme conditions. In contrast, we address the challenge of sustained balancing, requiring precise whole-body coordination and strong disturbance resilience.

\textbf{Learning-based Humanoid Control.} 
Recent years have seen rapid progress in learning-based methods for humanoid control. These approaches have demonstrated impressive success in humanoid locomotion~\cite{nakanishi2004learning,calandra2016bayesian,li2019using,li2021reinforcement,gu2024advancing,liao2024berkeley,radosavovic2024humanoid,radosavovic2024real,zhang2024whole,chen2024learning}. More recent work expands beyond basic walking to include agile and expressive behaviors such as running~\cite{crowley2023optimizing, li2024reinforcement}, jumping~\cite{li2023robust,he2025asap}, dancing~\cite{cheng2024expressive,ji2024exbody2}, parkour~\cite{zhuang2024humanoid,long2024learning}, and loco-manipulation~\cite{he2024learning,fu2024humanplus,he2024omnih2o,ben2025homie}. Despite achieving diverse whole-body motion, most focus on dynamic stabilization and do not address the precise balance control required for quasi-static poses. In contrast, our work introduces a balance-centric learning framework that emphasizes sustained stability in extreme configurations such as single-leg support.

\textbf{Sim-to-Real Transfer in Robot Learning.} 
Transferring policies from simulation to the real world remains a fundamental challenge. Common approaches include system identification, which calibrates simulation parameters using real-world data~\cite{khosla1985parameter,gautier1988identification,zhu2017fast,tan2016simulation,kolev2015physically,yu2017preparing,yu2019sim, allevato2020tunenet}; real-to-sim feedback, which corrects simulator behavior by incorporating real-world observations or learned residuals ~\cite{liu2020real,wang2023real2sim2real,torne2024reconciling,he2025asap,lin2025sim}; and domain randomization, which enhances robustness by training over a distribution of randomized dynamics and sensory conditions~\cite{tobin2017domain, peng2018sim, tobin2018domain, mehta2020active,huber2024domain}. While these strategies have shown success in locomotion and manipulation, their effectiveness in humanoid quasi-static balance remains underexplored, where even small sensor or contact inconsistencies can lead to instability. Building upon domain randomization, our method introduces balance-specific perturbations to improve real-world robustness.

\section{Learning Framework for Extreme Humanoid Balance}

\begin{figure*}
    \vspace{-3em}
    \centering
    \includegraphics[width=.95\textwidth]{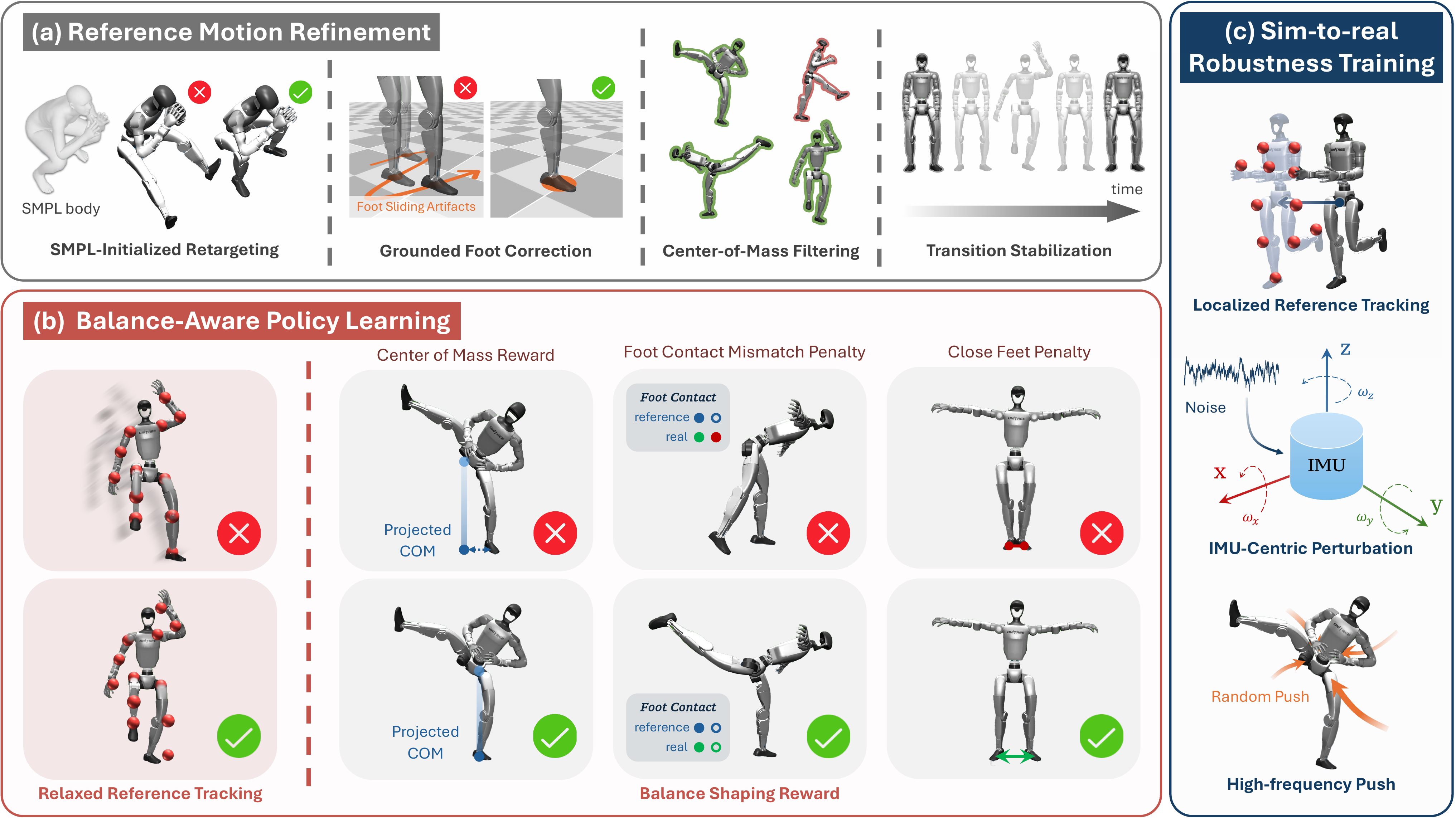}
    \caption{\small{\textbf{\method Overview.} To tackle the challenges of extreme balance tasks on humanoids, \method integrates three components: (a) a motion refinement process that improves the quality and feasibility of reference motions; (b) a balance-aware policy learning strategy that enables stable execution of challenging balance motions; and (c) a robustness training mechanism to improve sim-to-real consistency and deployment stability.}}
    \vspace{-1.9em}
    \label{fig:method}
\end{figure*}

As discussed in \Cref{sec:intro}, we identify three key challenges in learning extreme quasi-static balance tasks for humanoids. In this section, we first introduce the necessary background, then present a detailed description of the components of our proposed framework to address these challenges. An overview of the challenges and their corresponding solutions is illustrated in \Cref{fig:method}.

\subsection{Background}
\textbf{Problem Formulation.} Our balance learning framework adopts a tracking-based control paradigm, and we formulate the balance task as a goal-conditioned RL problem, modeled as a Markov Decision Process (MDP) $\mathcal{M} = \langle \mathcal{S}, \mathcal{A}, \mathcal{T}, \mathcal{R}, \gamma \rangle$, where $\mathcal{S}$ is the state space, $\mathcal{A}$ the action space, $\mathcal{T}$ the transition dynamics, $\mathcal{R}$ the reward function, and $\gamma$ the discount factor. Each state $\boldsymbol{s}_t \in \mathcal{S}$ includes the agent's proprioceptive observation $\boldsymbol{s}^{\text{p}}_t$ and a goal state $\boldsymbol{s}^{\text{g}}_t$ from the reference motion. The agent outputs actions $\boldsymbol{a}_t$ specifying desired joint angles, executed by a low-level proportional-derivative (PD) controller. The reward $r_t = \mathcal{R}(\boldsymbol{s}^{\text{p}}_t, \boldsymbol{s}^{\text{g}}_t, \boldsymbol{a}_t)$ encourages accurate tracking and stable control.

\textbf{Overall Pipeline.}
To enable tracking-based humanoid balance control, we first collect video clips and extract human poses using video-based motion capture algorithms such as WHAM~\cite{shin2024wham}, representing them in the SMPL format~\cite{SMPL:2015}. These poses are then retargeted to humanoid reference motions.
Based on the generated reference motions, we adopt a teacher-student learning paradigm~\cite{he2024omnih2o} to train balance policies. A teacher policy is first trained using Proximal Policy Optimization (PPO)~\cite{schulman2017proximal} with access to privileged observations. Then, a student policy is distilled via DAgger~\cite{ross2011reduction} using only onboard observations. All policies are trained in simulation~\cite{makoviychuk2021isaac}, and the final student policy is deployed on the real humanoid robot. Additional details are provided in \Cref{appendix:method_details}.

\subsection{Reference Motion Refinement}
Tracking performance in balance tasks is highly sensitive to the quality of reference motions, yet these often contain artifacts that hinder physical feasibility (see \Cref{fig:method} (a)). To mitigate this, we introduce a set of motion refinement strategies.

\textbf{SMPL-Initialized Retargeting.} Some prior humanoid retargeting approaches~\cite{he2024learning,he2024omnih2o} initialize joint angles to a zero pose and perform optimization by minimizing the positional differences between corresponding joints. However, in non-convex settings, such initialization can yields suboptimal results (see \Cref{fig:method}(a)), as the zero pose can be far from the optimal solution. To address this, we propose a more effective initialization strategy based on the human SMPL pose. Given the humanoid’s joint degrees of freedom form a subset of those in SMPL, we initialize each humanoid joint using the corresponding Euler angles from the SMPL pose. This SMPL initialization provides a starting point closer to the optimal solution, leading to faster convergence and improved accuracy.

\textbf{Grounded Foot Correction.} To further improve reference motion quality after retargeting, we introduce a data post-processing step to address foot instability caused by motion capture noise and retargeting errors. We assume that in the original human motion, the support foot should remain stationary without slipping. Based on this assumption, during single-legged phases, we adjust the global root position while keeping all local joint angles unchanged, ensuring that the grounded foot remains static across consecutive frames. This correction enhances foot stability and mitigates unrealistic foot-sliding artifacts introduced by noisy pose estimation and imperfect retargeting.

\textbf{Center-of-Mass Filtering.} Due to motion capture errors and human-humanoid mass mismatch, reference motions may exhibit large center-of-mass (COM) shifts, leading to physically infeasible trajectories, especially during single-leg phases. To address this, we compute the COM from the URDF-defined~\cite{ros_urdf} body masses and positions, and discard trajectories where the ground-projected COM deviates from the center of the support foot by more than $0.2\,\mathrm{m}$, ensuring feasible references.

\textbf{Transition Stabilization.}
Challenging balance motions are often sensitive to the initial pose, and even slight instability in the double-support stance prior to execution can adversely affect performance. To address this issue, we propose a simple yet effective strategy: extending the double-support phase before and after the balance phase. Specifically, we duplicate the first and last frames of the reference motion so that their total duration equals the balance phase. This not only increases the proportion of training time spent in stable double-foot stance—facilitating policy learning of standing balance—but also gives the humanoid more time to settle before transitioning into the extreme balance motion during deployment, ensuring a more reliable transition into the target pose.

\subsection{Balance-Aware Policy Learning}
Even with refined reference motions, humanoids face inherent challenges in maintaining balance due to morphological mismatch and the lack of structured guidance for balance behaviors. We overcome these issues through relaxed reference tracking and balance shaping rewards.

\textbf{Relaxed Reference Tracking.}
Due to structural differences and the resulting center of mass misalignment, closely tracking human motions on a humanoid often leads to instability. To mitigate this, we leverage the exploratory nature of RL and allow the policy to make fine-grained adjustments to the center of mass. Specifically, we relax the tracking objective by setting a relatively large tolerance ($\sigma = 0.6\,\mathrm{m}$) in the reward function (see \Cref{appendix:method_details/reward} for details), enabling the policy to deviate from the reference when strict adherence would compromise balance. This flexibility promotes the emergence of more stable behaviors, facilitating successful execution of extreme balance tasks.

\textbf{Balance Shaping Rewards.}
Merely relaxing the tracking constraint does not guarantee that the policy will learn physically feasible behaviors. Without structured guidance, reinforcement learning can converge to suboptimal solutions that violate task intent or physical plausibility. To handle this challenge, we design a set of shaping rewards. (i) \textbf{Center of mass (COM) reward} encourages the vertical projection of the COM to remain within the support polygon, thereby helping the policy learn to adjust its pose to satisfy balance constraints. (ii) \textbf{Foot contact mismatch penalty} categorizes the landing state of each foot as either in contact with the ground or not, and penalizes discrepancies between the humanoid's and the reference's contact states. For example, it discourages the non-supporting foot from making unintended ground contact during single-leg balancing. While such contact may offer momentary stability, it violates task constraints and compromises the integrity of the intended balance behavior. (iii) \textbf{Close feet penalty} prevents the feet from getting too close to one another, reducing the risk of inter-foot collisions and encouraging more stable lower-body poses. Collectively, these shaping rewards promote the emergence of balance-aware, physically plausible motion policies. More reward details are described in \Cref{appendix:method_details/reward}.

\subsection{Sim-to-Real Robustness Training}
Sensor noise—particularly from IMU and visual-inertial odometry (VIO) systems—and unmodeled real-world dynamics pose significant challenges for sim-to-real transfer in balance control. To enhance real-world robustness, we adopt localized reference tracking and IMU-centric observation perturbation to mitigate the issues caused by VIO and IMU noise, respectively, and apply high-frequency push disturbances to improve resilience against simulation modeling inaccuracies.

\textbf{Localized Reference Tracking Training.} To address the noise issues in VIO, we discard odometry information during both student policy training and deployment, and align the reference root with the humanoid’s current root pose, expressing all tracking targets in the local coordinate frame. Prior work, such as ExBody2~\cite{ji2024exbody2}, discards odometry only at deployment but relies on global tracking during training, leading to a mismatch that prevents the policy from accurately perceiving its own motion. As a result, the robot often fails to correct balance loss, persistently falling or jumping in a particular direction. In contrast, our approach maintains consistency between training and deployment while avoiding the adverse effects of VIO noise.

\label{subsec:Sim-to-Real Robustness Training}
\textbf{IMU-Centric Observation Perturbation.} Prior approaches~\cite{he2024learning,he2024omnih2o} inject uniform noise into observations to improve robustness, but this fails to capture the specific characteristics of IMU noise. Since the IMU-provided root orientation defines the local coordinate frame, many observation quantities—such as projected gravity, localized angular velocity, and localized reference targets—are intrinsically coupled. Simply adding independent uniform noise overlooks these dependencies. Moreover, real-world IMU noise exhibits significant temporal correlation. To address these issues, we perturb the observed root orientation—represented in Euler angles—with Ornstein-Uhlenbeck (OU) noise~\cite{uhlenbeck1930theory}, a temporally correlated stochastic process, during student training. All observations are then computed based on the perturbed orientation observation, ensuring that the resulting observation noise reflects both the temporal dynamics and the structural dependencies induced by IMU errors, thereby yielding a more realistic simulation of real-world sensor behavior.

\label{sec:method_sim2real}
\textbf{High-Frequency Push Disturbance.} Tracking-based policies often fail during real-world deployment of single-leg tasks, as minor initial oscillations can progressively amplify due to unmodeled dynamics. To better approximate this failure mode, we apply random external pushes during teacher policy training by injecting small, high-frequency velocity offsets into the root (push every 1s at up to 0.5 m/s). This strategy effectively incorporates real-world instability into simulation, significantly enhancing sim-to-real transferability and disturbance robustness. In contrast, prior work~\cite{he2024omnih2o} introduced low-frequency, high-magnitude pushes to train recovery from sudden external forces. However, such large perturbations are unsuitable for single-leg balance, where the feasible region is extremely narrow, and fail to capture the subtle instability dynamics critical for maintaining balance.

\section{Experiments}

Our experiments aim to answer the following questions: (1) How well does \method perform on extreme balance tasks compared to prior tracking-based approaches? (2) What are the contributions of each key component in \method's design to its overall performance? (3) Can \method transfer successfully to the real world, and how robust is it to external perturbations?

\subsection{Experiments Setup}
\label{sec:Experiments Setup}
\textbf{Environment and Tasks.}
We conduct our experiments on the Unitree G1 humanoid robot, evaluating \method across a set of balance tasks with varying difficulty levels (visualized in~\Cref{fig:teaser}). Simulation experiments are performed in the IsaacGym environment~\cite{makoviychuk2021isaac}.  To better simulate real-world jitter and sensor noise, we introduce two perturbations during testing: (i) random external pushes every 1s by perturbing root velocity up to 0.1 m/s, and (ii) IMU noise via Ornstein-Uhlenbeck (OU) noise~\cite{uhlenbeck1930theory} added to the root orientation in Euler angles. Each policy is evaluated over 100 episodes under these perturbed simulation conditions. For real robot setup, please refer to \Cref{appendix:real_robot_setup}.

\textbf{Metrics.} We design a set of metrics to comprehensively evaluate policy performance on balance tasks, organized into three categories: (1) \textbf{Task Completion.} \textit{Contact Mismatch (frame)} counts frames where foot contact states are incorrect (e.g., the non-supporting foot touches the ground during single-leg balancing). \textit{Success Rate (\%)} is the percentage of episodes where the humanoid maintains balance without (i) falling, (ii) foot contact mismatch, or (iii) an average tracking error exceeding 0.5 meters. (2) \textbf{Stability.} \textit{Slippage (mm/s)} measures the support foot’s ground-relative velocity, where higher values indicate unstable foot contact; \textit{Air (frame)} counts frames where both feet are airborne, typically indicating a loss of ground contact due to instability; \textit{Action Rate (rad/frame)} measures the action change magnitude between consecutive steps, where higher rates may suggest abrupt, unstable control behaviors. (3) \textbf{Tracking Error.} We report average global errors in keypoint position \textit{$E_{\text{pos}}$ (mm)}, velocity \textit{$E_{\text{vel}}$ (mm/frame)}, and acceleration \textit{$E_{\text{acc}}$ $\text{(mm/frame}^{2})$}. In real-world experiments, due to the absence of odometry, we instead compute local errors relative to the robot base frame, denoted as \textit{$E_{\text{pos-l}}$}, \textit{$E_{\text{vel-l}}$}, and \textit{$E_{\text{acc-l}}$}.

\textbf{Baselines.} 
To evaluate the performance of \method compared to standard tracking-based approaches, we consider the following baselines: (1) \textbf{H2O}~\cite{he2024learning}: a tracking-based humanoid control framework that retargets human motion data to the humanoid and trains a RL policy to track the reference motion. (2) \textbf{OmniH2O}~\cite{he2024omnih2o}: an extension of H2O that introduces a teacher-student learning paradigm, where a teacher policy is trained with privileged information using RL, and a student policy is distilled from it via DAgger~\cite{ross2011reduction} with only deployment-accessible observations. For a fair comparison, the baselines are adapted to our localized tracking framework and trained from scratch using the same balance motion data as \method, tracking the same set of keypoints.

\subsection{Simulation Results}

\begin{table*}[t]
\vspace{-10mm}
\centering
\resizebox{\linewidth}{!}{%
\begingroup
\setlength{\tabcolsep}{5pt}
\renewcommand{\arraystretch}{0.8}
\begin{threeparttable}
\begin{tabular}{lrrrrrrrrrrrrrrrr}
\toprule
\multicolumn{1}{c}{} & \multicolumn{8}{c}{\textbf{Swallow Balance}} & \multicolumn{8}{c}{\textbf{Bruce Lee's Kick}} \\
\cmidrule[\heavyrulewidth](r){2-9} \cmidrule[\heavyrulewidth](r){10-17}
\multicolumn{1}{c}{} & \multicolumn{2}{c}{Completion} & \multicolumn{3}{c}{Stability} & \multicolumn{3}{c}{Tracking Error} & \multicolumn{2}{c}{Completion} & \multicolumn{3}{c}{Stability} & \multicolumn{3}{c}{Tracking Error} \\
\cmidrule(r){1-1}\cmidrule(r){2-3} \cmidrule(r){4-6} \cmidrule(r){7-9} \cmidrule(r){10-11} \cmidrule(r){12-14} \cmidrule(r){15-17}
\textbf{Method} & $\text{Succ}\tnote{1} \uparrow$ & $\text{Cont}\tnote{2} \downarrow$ &  $\text{Slip}\tnote{3} \downarrow$  & $\text{Air} \downarrow$ & $\text{Act}\tnote{4} \downarrow$ &  $E_\text{pos} \downarrow$ &  $E_{\text{vel}} \downarrow$  & $E_{\text{acc}} \downarrow$ & $\text{Succ} \uparrow$ & $\text{Cont} \downarrow$ &  $\text{Slip} \downarrow$  & $\text{Air} \downarrow$ & $\text{Act} \downarrow$ &  $E_\text{pos} \downarrow$ &  $E_{\text{vel}} \downarrow$  & $E_{\text{acc}} \downarrow$ \\
\cmidrule(r){1-1}\cmidrule(r){2-3} \cmidrule(r){4-6} \cmidrule(r){7-9} \cmidrule(r){10-11} \cmidrule(r){12-14} \cmidrule(r){15-17}
H2O~\cite{he2024learning} & 0 & 181.85 & 203.24 & 2.29 & 8.02 & 572.82 & 8.18 & 4.39 & 4 & 4.57 & 368.65 & 17.11 & 9.53 & 328.75 & 7.79 & 5.13 \\ 
OmniH2O~\cite{he2024omnih2o} & 0 & 237.09 & 149.48 & 2.02 & 5.87 & 155.45 & 4.05 & 2.54 & 3 & 15.54 & 191.67 & 9.13 & 2.47 & 116.21 & 6.27 & 3.41 \\  
\method (ours)& \textbf{100} & \textbf{0.00} & \textbf{80.81} & \textbf{0.89} & \textbf{0.51} & \textbf{83.15} & \textbf{3.50} & \textbf{1.89}  & \textbf{100} & \textbf{0.00} & \textbf{76.44} & \textbf{1.28} & \textbf{0.46} & \textbf{67.18} & \textbf{3.31} & \textbf{2.33} \\
\cmidrule(r){1-1}\cmidrule(r){2-3} \cmidrule(r){4-6} \cmidrule(r){7-9} \cmidrule(r){10-11} \cmidrule(r){12-14} \cmidrule(r){15-17}
\rowcolor{lightgray} 
\multicolumn{17}{l}{\textbf{(a) Ablation on Relaxed Tracking}} \\
\cmidrule(r){1-1}\cmidrule(r){2-3} \cmidrule(r){4-6} \cmidrule(r){7-9} \cmidrule(r){10-11} \cmidrule(r){12-14} \cmidrule(r){15-17}
\method-track-sigma-0.15m & 97 & 0.01 & 103.89 & 3.33 & 0.81 & \textbf{67.74} & \textbf{3.20} & 1.99 & \textbf{100} & \textbf{0.00} & 96.73 & 2.59 & 0.80 & 69.43 & \textbf{3.29} & 2.43\\
\method-track-sigma-0.3m & 99 & \textbf{0.00} & 107.12 & 3.76 & 0.63 & 89.50 & 3.29 & 1.95 & 99 & \textbf{0.00} & 104.67 & 5.46 & 0.58 & 89.09 & 3.48 & 2.45\\ 
\method-track-sigma-1.2m & 73 & 0.41 & \textbf{54.64} & \textbf{0.14} & \textbf{0.47} & 223.96 & 6.52 & 2.11 & 99 & 0.03 & \textbf{72.34} & \textbf{1.02} & \textbf{0.43} & 80.61 & 3.37 & \textbf{2.33} \\ 
\method-track-sigma-0.6m (ours) & \textbf{100} & \textbf{0.00} & 80.81 & 0.89 & 0.51 & 83.15 & 3.50 & \textbf{1.89} & \textbf{100} & \textbf{0.00} & 76.44 & 1.28 & 0.46 & \textbf{67.18} & 3.31 & \textbf{2.33} \\ 
\cmidrule(r){1-1}\cmidrule(r){2-3} \cmidrule(r){4-6} \cmidrule(r){7-9} \cmidrule(r){10-11} \cmidrule(r){12-14} \cmidrule(r){15-17}
\rowcolor{lightgray} 
\multicolumn{17}{l}{\textbf{(b) Ablation on Balance Shaping Rewards}} \\
\cmidrule(r){1-1}\cmidrule(r){2-3} \cmidrule(r){4-6} \cmidrule(r){7-9} \cmidrule(r){10-11} \cmidrule(r){12-14} \cmidrule(r){15-17}
\method-w/o-COM-reward & 99 & \textbf{0.00} & 91.29 & 1.09 & 0.53 & 91.74 & 3.58 & \textbf{1.89} & 98 & \textbf{0.00} & 76.86 & 3.43 & \textbf{0.40} & 67.71 & \textbf{3.18} & \textbf{2.28} \\ 
\method-w/o-contact-penalty & 74 & 0.49 & 93.39 & 1.53 & 0.58 & 104.14 & 3.96 & 2.03 & 62 & 0.87 & 78.43 & 2.01 & 0.56 & 67.66 & 3.31 & 2.33 \\ 
\method-w/o-close-feet-penalty & 96 & 0.06 & 102.76 & \textbf{0.75} & 0.67 & 123.00 & 3.97 & 2.02 & \textbf{100} & \textbf{0.00} & 83.47 & 1.77 & 0.53 & 81.42 & 3.52 & 2.37 \\
\method & \textbf{100} & \textbf{0.00} & \textbf{80.81} & 0.89 & \textbf{0.51} & \textbf{83.15} & \textbf{3.50} & \textbf{1.89} & \textbf{100} & \textbf{0.00} & \textbf{76.44} & \textbf{1.28} & 0.46 & \textbf{67.18} & 3.31 & 2.33 \\
\cmidrule(r){1-1}\cmidrule(r){2-3} \cmidrule(r){4-6} \cmidrule(r){7-9} \cmidrule(r){10-11} \cmidrule(r){12-14} \cmidrule(r){15-17}
\rowcolor{lightgray} 
\multicolumn{17}{l}{\textbf{(c) Ablation on Sim-to-Real Robustness Training}} \\
\cmidrule(r){1-1}\cmidrule(r){2-3} \cmidrule(r){4-6} \cmidrule(r){7-9} \cmidrule(r){10-11} \cmidrule(r){12-14} \cmidrule(r){15-17}
\method-w/o-localized-tracking & 92 & 5.65 & 152.62 & 11.14 & 1.75 & 311.56 & 5.24 & 2.55 & 99 & 0.01 & 116.86 & 2.74 & 1.24 & 187.27 & 4.69 & 2.93 \\
\method-w/o-imu-noise & 92 & 2.13 & 233.40 & 17.10 & 2.53 & 253.77 & 6.56 & 3.13 & 93 & 0.62 & 287.32 & 14.53 & 5.14 & 268.26 & 7.14 & 4.19\\ 
\method-w/o-push & 90 & 3.65 & 102.50 & 4.75 & 0.74 & 98.06 & 3.75 & 1.97 & 89 & 1.19 & 156.69 & 7.67 & 6.40 & 134.01 & 4.63 & 3.04\\ 
\method-push (5s interval, 1\,m/s) & 97 & 0.33 & 99.09 & 2.78 & 0.83 & 141.74 & \textbf{3.49} & 1.91 & \textbf{100} & \textbf{0.00} & \textbf{74.37} & 1.51 & \textbf{0.43} & 69.10 & 3.38 & 2.34 \\ 
\method & \textbf{100} & \textbf{0.00} & \textbf{80.81} & \textbf{0.89} & \textbf{0.51} & \textbf{83.15} & 3.50 & \textbf{1.89} & \textbf{100} & \textbf{0.00} & 76.44 & \textbf{1.28} & 0.46 & \textbf{67.18} & \textbf{3.31} & \textbf{2.33}\\
\bottomrule 
\end{tabular}

\begin{tablenotes}
Abbreviation for 
\item [1] \textit{Success Rate}
\item [2] \textit{Contact Mismatch}
\item [3] \textit{Slippage}
\item [4] \textit{Action Rate}
\end{tablenotes}
\vspace{-3mm}
\end{threeparttable}
\endgroup}
\caption{
\small{\textbf{Simulation Results.} We compare \method against baselines and ablations. The results demonstrate that \method successfully completes extreme balance tasks, whereas baselines consistently fail. Ablation studies further highlight the critical contributions of each component of \method to the overall balance performance.}
}
\label{tab:sim_ablation_baseline}
\vspace{-5mm}
\end{table*}

As shown in \Cref{tab:sim_ablation_baseline}, we present the quantitative results of \method, the baselines, and the ablations on two challenging tasks, with additional results provided in \Cref{appendix:exp_details}.

\textbf{\method and Baselines Performance.}
The results demonstrate that \method is the only method that completes these extreme balance tasks with a 100\% success rate, while the baselines almost always fail due to large contact mismatches—specifically, unintended ground contacts by the non-supporting foot, which should remain airborne during single-leg balancing. Moreover, \method exhibits smaller ground slippage, shorter airborne durations, and lower action variability, indicating stronger stability and smoother motion execution. In addition, \method achieves lower tracking errors, suggesting more accurate task completion.

\textbf{Ablations on Relaxed Tracking.} We experiment with different tracking tolerance $\sigma$ values. The results show that a smaller $\sigma$ reduces the tracking error but simultaneously increases slippage, airborne time, and action rate, indicating degraded overall stability. In contrast, excessively large $\sigma$ values lead to improved stability but at the cost of significantly higher tracking errors, more frequent contact mismatches, and lower task success rates. Based on these observations, we select a moderately large tolerance of $\sigma = 0.6\,\mathrm{m}$, which achieves a favorable balance between tracking fidelity and policy stability compared to prior tracking-based methods.

\textbf{Ablations on Shaping Rewards.}
Removing the COM reward leads to increased slippage and airborne time, underscoring its critical role in stabilizing the center of mass to maintain balance. Eliminating the foot contact mismatch penalty results in a significant rise in contact mismatches and a notable drop in success rates, demonstrating its importance for ensuring successful single-foot balancing. Additionally, removing the close-feet penalty degrades performance, primarily because the humanoid's feet can come excessively close, which reduces overall standing stability.

\textbf{Ablations on Robustness Training.} First, replacing localized tracking with global tracking during training—while still deploying with localized tracking, as in ExBody2~\cite{ji2024exbody2}—introduces a mismatch between training and deployment. This significantly degrades performance on tasks like \texttt{Swallow Balance}, where successful execution depends on the precise completion of preceding motions. Second, removing IMU noise injection leads to performance deterioration across all metrics, indicating policies not exposed to sensor noise during training are highly sensitive to deployment errors. 
Third, omitting high-frequency push perturbations increases contact mismatches, lowers success rates, and degrades stability, indicating that perturbation exposure during training is critical for successful task execution and overall stability. 
Finally, replacing high-frequency pushes with low-frequency ones (push every 5s at up to 1 m/s), as in prior tracking-based methods~\cite{he2024omnih2o}, also degrades stability, especially on tasks with narrow feasible balance regions like \texttt{Swallow Balance}. 
\begin{wrapfigure}{r}{0.34\textwidth}
\vspace{-5mm}
  \begin{center}
    \includegraphics[width=0.34\textwidth]{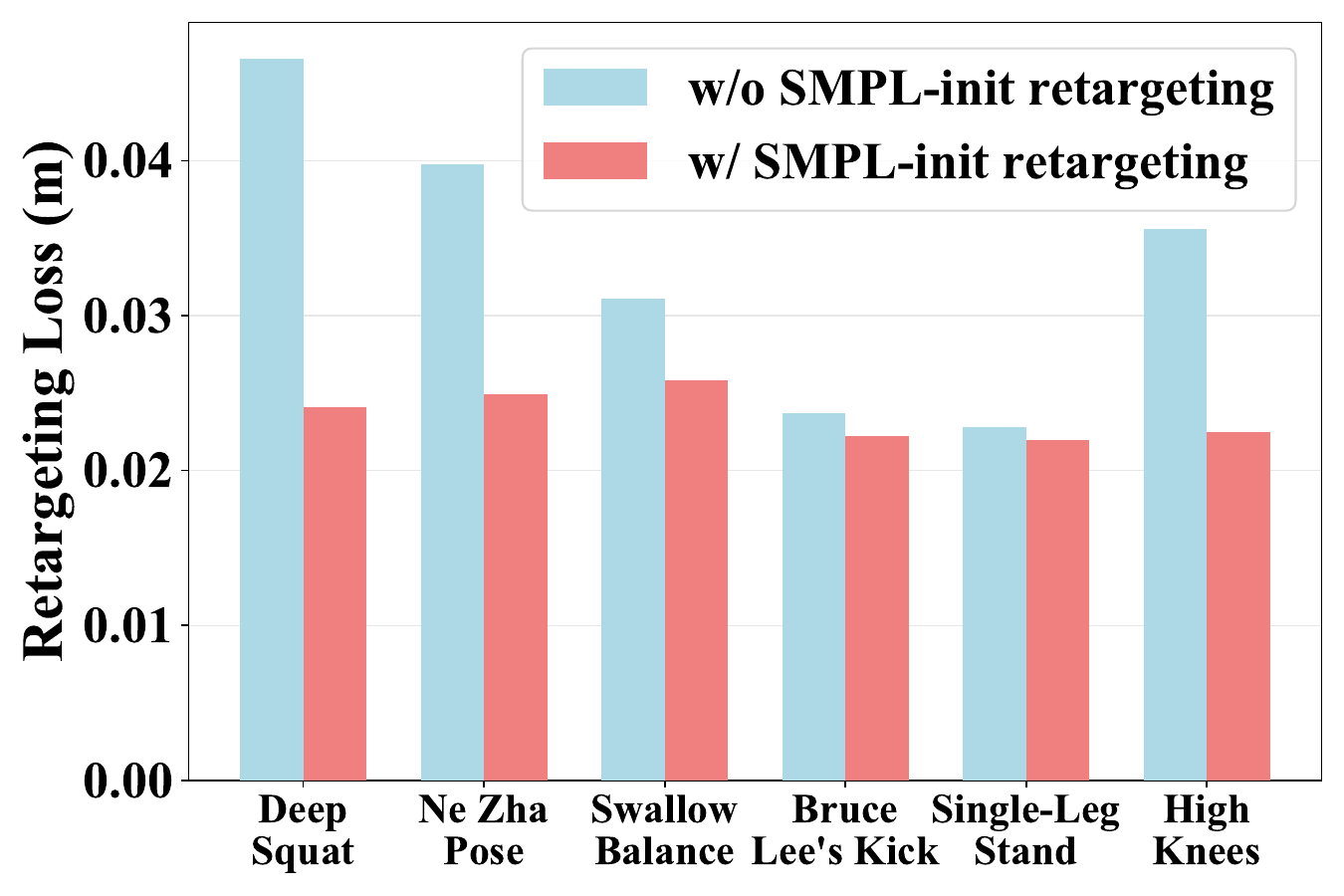}
  \end{center}
  \vspace{-3mm}
  \caption{\small{Retargeting Comparison.}}
  \label{fig:retarget_loss}
  \vspace{-5mm}
\end{wrapfigure}
This is likely because infrequent large perturbations are too hard for the humanoid to withstand during balance tasks.


\textbf{Retargeting Results.} To assess the impact of SMPL-initialization on retargeting, we compare the retargeting loss after 500 optimization steps between solutions optimized with and without it. As shown in \Cref{fig:retarget_loss}, solutions with SMPL-initialization consistently achieve lower losses across all tasks, with notably large reductions in \texttt{Deep Squat}.

\subsection{Real-World Results}

\begin{wraptable}{r}{0.3\linewidth}
\vspace{-4mm}
\centering
\resizebox{\linewidth}{!}{
\begingroup
\setlength{\tabcolsep}{2pt}
\renewcommand{\arraystretch}{0.8}
\begin{tabular}{lcccc}
\toprule
\textbf{Method} & $\text{Succ} \uparrow$ & $E_\text{pos-l} \downarrow $ &  $E_{\text{vel-l}} \downarrow$  & $E_{\text{acc-l}} \downarrow$   \\
\cmidrule(r){1-1}\cmidrule(r){2-5}
\rowcolor{lightgray} 
\multicolumn{5}{l}{\textbf{(a) Swallow Balance}} \\
\cmidrule(r){1-1}\cmidrule(r){2-5}
        OmniH2O & 0/5 & 119.73 & 2.20 & 1.86 \\ 
        \method & \textbf{4/5} & \textbf{38.31} & \textbf{1.73} & \textbf{1.13} \\
\cmidrule(r){1-1}\cmidrule(r){2-5}
\rowcolor{lightgray} 
\multicolumn{5}{l}{\textbf{(b) Bruce Lee's Kick}} \\
\cmidrule(r){1-1}\cmidrule(r){2-5}
        OmniH2O & 0/5 & 80.69 & 5.80 & 4.80 \\
        \method & \textbf{5/5} & \textbf{27.87} & \textbf{1.58} & \textbf{1.14} \\
\cmidrule(r){1-1}\cmidrule(r){2-5}
\rowcolor{lightgray} 
\multicolumn{5}{l}{\textbf{(c) Ne Zha Pose}} \\
\cmidrule(r){1-1}\cmidrule(r){2-5}
        OmniH2O & 0/5 & 50.48 & 1.29 & 1.37 \\
        \method & \textbf{5/5} & \textbf{30.91} & \textbf{0.73} & \textbf{0.37} \\ 
\cmidrule(r){1-1}\cmidrule(r){2-5}
\rowcolor{lightgray} 
\multicolumn{5}{l}{\textbf{(d) Single-leg Stand}} \\
\cmidrule(r){1-1}\cmidrule(r){2-5}
        OmniH2O & 0/5 & 32.10 & \textbf{0.49} & \textbf{0.25} \\
        \method & \textbf{5/5} & \textbf{29.58} & 0.96 & 0.47 \\ 
\cmidrule(r){1-1}\cmidrule(r){2-5}
\rowcolor{lightgray} 
\multicolumn{5}{l}{\textbf{(e) Deep Squat}} \\
\cmidrule(r){1-1}\cmidrule(r){2-5}
        OmniH2O & 4/5 & 42.97 & 3.21 & 4.21 \\
        \method & \textbf{5/5} & \textbf{29.90} & \textbf{2.08} & \textbf{1.08} \\ 
\bottomrule 
\end{tabular}
\endgroup}
\caption{\small{Real-World Results.}}
\vspace{-3mm}
\label{tab:real}
\end{wraptable}

\textbf{Balance Performance.} We evaluate the performance of \method and the baseline OmniH2O on real-world balance tasks. \Cref{fig:teaser} visualizes real-world executions of \method, and ~\Cref{tab:real} quantitatively compares \method and OmniH2O across evaluation metrics.  Videos are available on the project website. The results show that \method successfully completes challenging balance tasks in the real world, including \texttt{Swallow Balance} and \texttt{Bruce Lee’s Kick}, which are difficult even for humans. The humanoid holds these extreme poses with stability and fluidity, maintaining steady balance and making only minor foot adjustments when necessary. These results highlight the strong balance capabilities of \method. In contrast, the baseline struggles to complete the tasks: for example, it fails to maintain balance and falls during \texttt{Swallow Balance}, exhibits excessive shaking during \texttt{Bruce Lee’s Kick}, and abandons leg lifting completely in \texttt{Single-Leg Stand} and \texttt{Ne Zha Pose} to reduce the risk of falling.

\begin{wrapfigure}{r}{0.31\textwidth}
\vspace{-8mm}
  \begin{center}
    \includegraphics[width=0.3\textwidth]{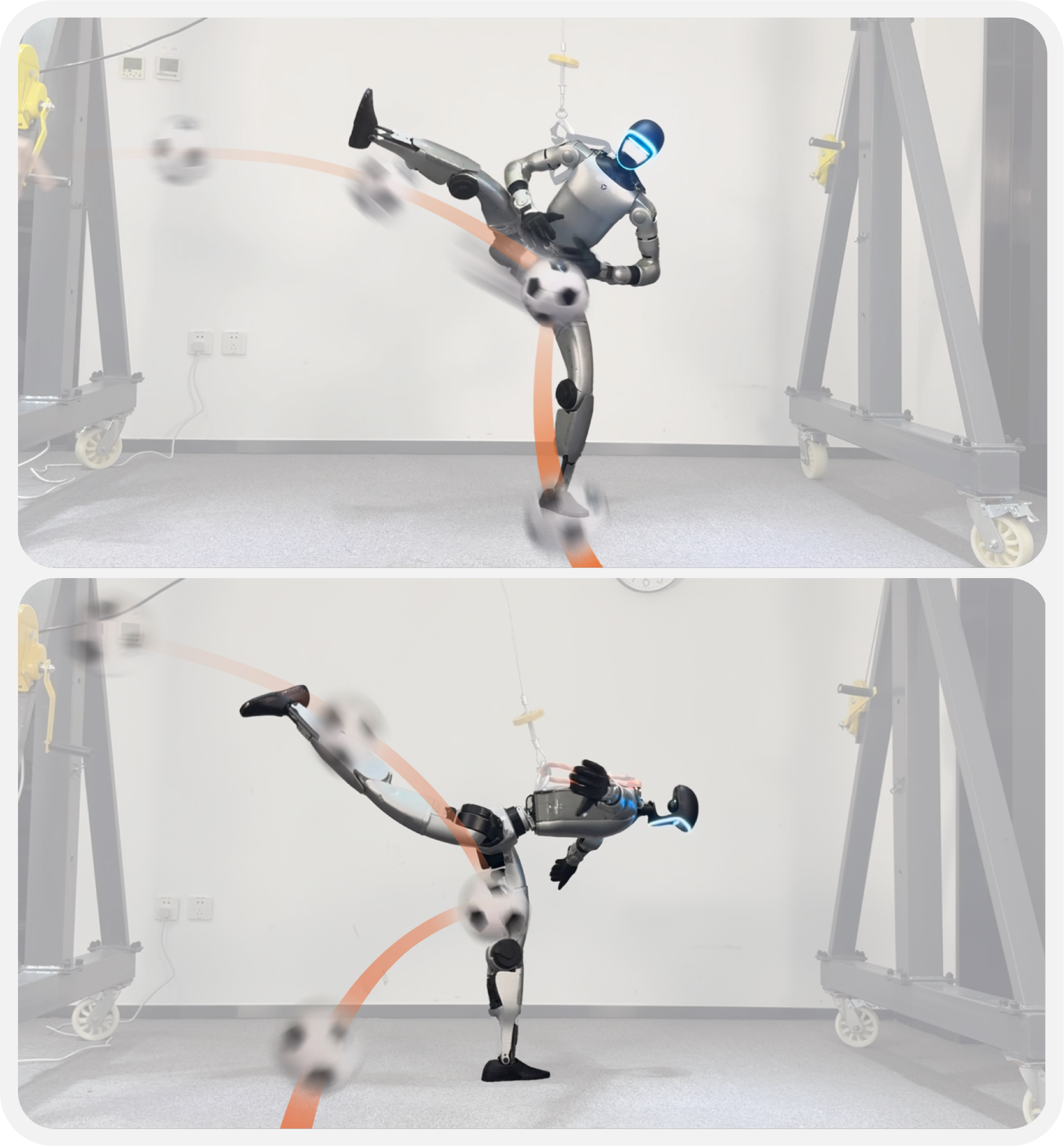}
  \end{center}
  \vspace{-2mm}
  \caption{\small External Perturbations.}
  \label{fig:perturbation}
  \vspace{-5mm}
\end{wrapfigure}

\textbf{Robustness Evaluation.} We evaluate the robustness of \method across two aspects. (1) \textbf{External Perturbations.} As illustrated in \Cref{fig:perturbation}, we apply external disturbances by striking the humanoid with a forcefully kicked soccer ball during balance tasks. Despite significant disruptions to the robot's balance, \method rapidly reacts and recovers with minimal corrective motion, returning to a stable state within a short period, demonstrating strong disturbance tolerance. (2) \textbf{Long-Horizon Task Execution.}  We conduct repeated trials of the \texttt{Bruce Lee’s Kick} task without resetting the humanoid between trials. \method successfully completes 10 consecutive executions in a single take without any failures or external intervention. This demonstrates \method's strong reliability, balance consistency, and control stability in real-world deployment. These two evaluations highlight the robustness of our framework and validate the effectiveness of its design.

\section{Conclusion}

We present \method, a unified learning-based framework for humanoid control in extreme balance tasks. By systematically addressing challenges such as reference motion inaccuracies, balance policy learning difficulties, and the sim-to-real gap, \method enables humanoid robots to stably execute challenging balance poses that baseline methods consistently fail to complete. It further demonstrates strong robustness to disturbances and consistency over long-horizon deployments.

\section{Limitations}
One limitation of our method is that certain components are specifically designed for balance tasks and rely on task-specific assumptions. As a result, they may not be directly applicable to other task categories, such as jumping or parkour. Moreover, although the trained policies are capable of accomplishing complex balancing behaviors, they exhibit limited generalization: adapting to novel and substantially different motions typically necessitates retraining. Developing policies capable of acquiring versatile motor skills that can be reliably deployed in the real world remains an important research direction.


\clearpage
\acknowledgments{We thank Jiacheng You, Haoyang Weng, and Bike Zhang for their helpful discussions, and Junming Zhao and Sicong Dai for their assistance with the real-world experiments.
This work is supported by the National Key R\&D Program of China (2022ZD0161700), National Natural Science Foundation of China (62176135, 12201341), Shanghai Qi Zhi Institute Innovation Program SQZ202306 and the Tsinghua University Dushi Program.}


\bibliography{bib}  

\newpage
\appendix

\section{Real Robot Setup}
\label{appendix:real_robot_setup}
We conduct our experiments on the Unitree G1 humanoid robot, which features 29 degrees of freedom (DoF), including two 7-DoF arms, two 6-DoF legs, and a 3-DoF waist. For real-world deployment, we use the robot’s onboard IMU to obtain root orientation and angular velocity, and joint encoders to obtain joint positions and velocities. The control policy receives keypoint tracking targets and proprioceptive information as input, computes the desired joint positions for each actuator, and sends commands to the robot’s low-level interface. Policy inference is executed in real time on the onboard NVIDIA Jetson Orin NX, with a control frequency of 50 Hz. Observations, including keypoint tracking information and proprioceptive data, are transmitted to the control policy via DDS~\cite{unitree2025dds}, using the \texttt{unitree\_sdk2\_python} implementation~\cite{unitree_sdk2_python}.
\section{\method Details}
\label{appendix:method_details}
\subsection{State Space Design} 
\label{appendix:method_details/state_space}
This subsection details the state space design for both the teacher and student policies in \method.

\textbf{Teacher Policy.} The teacher policy, trained via RL, has access to the full states required for reference tracking. \Cref{tab:teacher_state} presents the state space of the teacher policy.

\begin{table}[H]
\centering
\renewcommand{\arraystretch}{1.2}
\begin{tabular}{ c c }

\hline
\textbf{State term}  & \textbf{Dimensions}                              \\ 
\hline
Rigid body position & 87 \\
Rigid body rotation & 180 \\
Rigid body velocity & 90 \\
Rigid body angular velocity & 90 \\
Rigid body position difference & 90 \\
Rigid body rotation difference & 180 \\
Rigid body velocity difference & 90 \\
Rigid body angular velocity difference & 90 \\
Local reference rigid body position & 90 \\
Local reference rigid body rotation & 180 \\
Actions                     & {29} \\
\hline
\rowcolor{lightgray}
Total dim & {1196}\\
\hline
\end{tabular}
\vspace{3mm}
\caption{State space information of the teacher policy.}
\label{tab:teacher_state}
\end{table}

\textbf{Student Policy.} The student policy, trained using DAgger with a history of 25 steps, is restricted to deployment-accessible observations only. \Cref{tab:student_state} presents the state space of the student policy. For the student policy, we select a total of 12 tracking keypoints, corresponding to the left and right sides of the hips, knees, ankles, shoulders, elbows, and wrists.

\begin{table}[H]
\centering
\begin{tabular}{ c c }

\hline
\textbf{State term} & \textbf{Dimensions} \\ 
\hline
DoF position & {29}           \\
DoF velocity    & {29}        \\
Base angular velocity    & {3}       \\
Projected gravity             & {3}     \\
Localized reference keypoints position            & {36}      \\
Keypoints position difference   & {36}      \\
Keypoints velocity difference   & {36}      \\
Actions                     & {29} \\
\hline
\rowcolor{lightgray}
Single step total dim & {201}\\
\hline
History state term & Dimensions\\
\hline
DoF position & {29}           \\
DoF velocity    & {29}        \\
Base angular velocity    & {3}       \\
Projected gravity             & {3}     \\
Actions                     & {29} \\
\rowcolor{lightgray}
\hline
History single step total dim & {93}\\
\hline
\rowcolor{lightgray}
\hline
Total dim & {2526 (201 + 93$\times$25)}\\
\hline

\end{tabular}
\vspace{2mm}
\caption{State space information of the student policy.}
\label{tab:student_state}
\end{table}

\subsection{Rewards}
\label{appendix:method_details/reward}

\Cref{tab:appendix_reward} provides a summary of the detailed reward components.

\begin{table}[H]
\resizebox{\linewidth}{!}{%
\renewcommand{\arraystretch}{1.1}

\centering
\begin{threeparttable}
\begin{tabular}{  c  c  c  c }
\hline
\textbf{Term} & \textbf{Expression} & \textbf{Weight} & \textbf{Remarks}    \\ 
\hline
    &  Balance Shaping Rewards & \\
\hline
    Center of mass & $\exp(-\lVert \bm{p}_\mathrm{xy}^{\text{com}} - \bm{p}_\mathrm{xy}^{\text{lower-foot}} \rVert_2^2/ \sigma^2_\text{com}) \times \mathds{1}(\lVert \hat{\bm{p}}_\mathrm{z}^\text{l-foot}- \hat{\bm{p}}_\mathrm{z}^\text{r-foot} \rVert_2 > 0.05)$ & $160$ &  $\sigma_\text{com} = 0.1$\\
    Foot contact mismatch & $\bm{c}_\text{feet}\oplus \hat{\bm{c}}_\text{feet}$\tnote{1} & $-250$\\
    Close feet & $\max\{0.16 - \lVert \bm{p}^\text{l-foot}- \bm{p}^\text{r-foot}\rVert_2, 0\}$ & $-1000$\\
\hline
    &    Tracking Rewards  &           \\ 
\hline
Body position         &   $\exp (-\lVert \boldsymbol{p}_t-\hat{\boldsymbol{p}}_t\rVert_2^2 / \sigma^2_\text{pos}) $      & $30$ & $\sigma_\text{pos}=0.6 $    \\
Body rotation         &   $\exp(-\lVert \boldsymbol{\theta}_t \ominus \hat{\boldsymbol{\theta}}_t\rVert_2^2 / \sigma^2_\text{rot})$         & $20$  & $\sigma_\text{rot}=0.3$   \\
Body velocity         &    $\exp (-\lVert \boldsymbol{v}_t-\hat{\boldsymbol{v}}_t\rVert_2^2 / \sigma^2_\text{vel}) $        & $8$  & $\sigma_\text{vel}=3$    \\
Body angular velocity &   $\exp (-\lVert \boldsymbol{\omega}_t-\hat{\boldsymbol{\omega}}_t\rVert_2^2 / \sigma^2_\text{ang}) $         & $8$   & $\sigma_\text{ang}=10$   \\ 
DoF position        &   $\exp(-\lVert \boldsymbol{d}_t - \hat{\boldsymbol{d}}_{t}\rVert_2^2 / \sigma^2_\text{dpos})$         & $32$  & $\sigma_\text{dpos}=0.7$  \\
DoF velocity        &   $\exp(- \lVert \bm{\dot{d}}_t - \hat{\bm{\dot{d}}}_t \rVert_2^2 / \sigma^2_\text{dvel})$          & $16$  & $\sigma_\text{dvel}=10$  \\
\hline
               &     Penalty       &           \\ \hline
Torque limits        &     $ \mathds{1}(\boldsymbol{\tau}_{t} \notin [\boldsymbol{\tau}_{\min}, \boldsymbol{\tau}_{\max} ]) $          & $-0.5$   \\
DoF position limits      &   $ \mathds{1}({\boldsymbol{d}}_{t} \notin [\boldsymbol{d}_{\min}, \boldsymbol{d}_{\max} ])  $      & $-30$    \\
DoF velocity limits      &   $ \mathds{1}({\boldsymbol{\dot{d}}}_{t} \notin [\dot{\boldsymbol{d}}_{\min}, \dot{\boldsymbol{d}}_{\max} ])  $      & $-12$    \\
Termination           &     $\mathds{1}_\text{termination}$       & $-60$    \\ \hline
        &    Regularization       &           \\ \hline
Torque                & $\lVert \boldsymbol{\tau}_t \rVert$     & $-2.5\times10^{-5}$   \\
DoF velocity               &   $\lVert \boldsymbol{\dot{d}}_t \rVert_2^2$         & $-1\times10^{-3}$   \\
DoF acceleration               &   $\lVert \boldsymbol{\ddot{d}}_t \rVert_2$              & $-3\times10^{-6}$ \\
Action rate     &    $ \lVert \boldsymbol{a}_t - \boldsymbol{a}_{t-1} \rVert_2^2  $     & $-1.5$   \\
Feet air time       &      $T_\text{air}-0.25$~\cite{rudin2022learning}    & $250$     \\ 
Feet contact force   &     $\max\{\lVert F_\text{\text{feet}}\rVert_2 - 500, 0\}$       & $-0.2$   \\
Stumble               &     $\mathds{1}(F_\text{\text{feet}}^{xy} > 5\times F_\text{\text{feet}}^z)$       & $-3\times10^{-4}$    \\
Slippage              &    $\lVert \boldsymbol{v}_t^\text{\text{feet}} \rVert_2^2 \times \mathds{1} (F_\text{\text{feet}} \geq 1)$        & $-30$  \\
Feet orientation              &   $\lVert \boldsymbol{g}_\mathrm{z}^{\text{feet}} \rVert \times \mathds{1}(\bm{p}_\mathrm{z}^\text{feet} < 0.05)$      & $-62.5$   \\
In the air & $\mathds{1} (F_{\text{feet}}^{\text{left}}, F_{\text{feet}}^{\text{right}} < 1)$ & $-50$ \\
  \hline
\end{tabular}%
\begin{tablenotes} 
\item [1] $\bm{c}_\text{feet}$ represents the robot's feet contact with the ground, and $\hat{\bm{c}}_\text{feet}$ the reference's. Whether the robot's feet are in contact is determined by $F_{\text{feet}} \geq 1\,\mathrm{N}$. For the reference, both feet are considered grounded if their height difference is below 0.05m; otherwise, the lower foot is considered grounded.
\end{tablenotes}

\end{threeparttable}
}
\vspace{2mm}
\caption{Reward components and weights. Quantities with the hat symbol ($\hat{\cdot}$) represent reference motion variables, while unmarked terms refer to the humanoid's own state variables.
}
\label{tab:appendix_reward}
\end{table}

\subsection{Domain Randomization}
\label{appendix:method_details/domain_rand}

\Cref{tab:domain_rand} summarizes the domain randomization strategies used in \method, including high-frequency push disturbances designed to bridge the sim-to-real gap and improve balance robustness.
\begin{table}[H]
\centering
\renewcommand{\arraystretch}{1}
\begin{tabular}{ c c }
\hline
\textbf{Term}  & \textbf{Value} \\ 
\hline
\rowcolor{lightgray}
\multicolumn{2}{c}{High-Frequency Push Disturbance}  \\ \hline
Push robot         & $\text{interval}=1\,\mathrm{s}$, $ v_\mathrm{xy} \in \mathcal{U}(0, 0.5) \, \text{m/s}$ \\
\hline

\rowcolor{lightgray}
\multicolumn{2}{c}{Dynamics Randomization}  \\ 
\hline
Friction  coefficient & $\mathcal{U}(2.5, 3.5)$            \\
Torso COM offset    & $\mathcal{U}(-0.1, 0.1) \, \text{m}$           \\
Link mass    & $\mathcal{U}(0.7, 1.3) \times \text{default} \ \text{kg}$            \\
PD gains             & $\mathcal{U}(0.75, 1.25) \times \text{default}$          \\
Torque RFI~\cite{campanaro2023learning}         & $0.1 \times \text{torque limit}\ \text{N}\cdot\text{m}$  \\
Control delay      & $\mathcal{U}(20, 60)\text{ms}$           \\ 
Motion reference offset      & $\mathcal{U}([-0.02,0.02],[-0.02,0.02],[-0.1, 0.1])$\,m          \\ 
\hline
\end{tabular}
\vspace{2mm}
\caption{ Domain randomizations for \method.}
\label{tab:domain_rand}
\end{table}

\subsection{IMU Noise}
As illustrated in \Cref{subsec:Sim-to-Real Robustness Training}, we introduce Ornstein-Uhlenbeck (OU) noise ~\cite{uhlenbeck1930theory} to the IMU's Euler angles observation (in degree). OU noise is modeled by the following differential equation:

\vspace{-1mm}

$$
\frac{\mathrm{d}X_t}{\mathrm{d}t} = -\theta X_t + \sigma \epsilon_t
$$

where $X_t$ represents the OU noise, $\theta$ is the mean reversion rate, $\sigma$ is the noise intensity, and $\epsilon_t$ is a standard Gaussian noise term ($\epsilon_t \sim \mathcal{N}(0, 1)$) at each time step. The noise term introduces random fluctuations, while the mean reversion term prevents excessive drift. For our experiments, we set the parameters to $\theta = 25$ and $\sigma = 250$.

\subsection{Hyperparameters}
\Cref{tab:hyperparam} presents the hyperparameters used for training \method.
\begin{table}[H]
\centering
\begin{tabular}{l c} 
\hline
\textbf{Hyperparameters} & \textbf{Values} \\ \hline
Optimizer & Adam \\
$\beta_1, \beta_2$ & 0.9, 0.999 \\
Learning rate & $1 \times 10^{-3}$\\
Batch size & 64\\
Discount factor ($\gamma$) & 0.99\\
Clip param & 0.2 \\
Entropy coef & 0.005 \\
Max grad norm &  0.2\\
Value loss coef & 1 \\
Entropy coef & 0.005 \\
Init noise std (RL) & 1.0 \\
Init noise std (DAgger) & 0.001 \\
Num learning epochs & 5\\
MLP size & [512, 256, 128]\\  \hline
\end{tabular}
\vspace{2mm}
\caption{Hyperparameters.}
\label{tab:hyperparam}
\end{table}

\subsection{Foot Contact Labeling}
In the \textit{grounded foot correction} stage, the foot with the lower ankle height is considered the grounded foot. For the \textit{foot contact mismatch penalty}, please refer to \Cref{tab:appendix_reward} for the criteria used to determine which foot is considered grounded.
\section{Experiments Details}
\label{appendix:exp_details}

\subsection{Experiments Setup Details}
It is worth noting that, for a fair comparison, all baselines (OmniH2O and H2O) are trained from scratch using the same set of balance motion data as \method, and are tasked with tracking the same set of keypoints.

To better approximate real-world conditions, we apply the same domain randomization during both training and evaluation, except for the random external pushes. As described in \Cref{sec:method_sim2real} and \Cref{sec:Experiments Setup}, different push magnitudes are used for training and evaluation—larger magnitudes (0.5 m/s) are applied during training to ensure the policy learns robustness under stronger disturbances, while smaller perturbations (0.1 m/s) are used in evaluation to more closely reflect realistic deployment scenarios.

\subsection{Additional Results}

\Cref{tab:additional_exp} shows the performance of \method and baselines across additional three tasks. \method consistently outperforms the baselines in completion, stability, and tracking errors, demonstrating superior performance.

\begin{table}[H]
\centering
\resizebox{\linewidth}{!}{%
\begingroup
\setlength{\tabcolsep}{8pt}
\renewcommand{\arraystretch}{0.8}
\begin{threeparttable}
\begin{tabular}{lrrrrrrrr}
\toprule
\multicolumn{1}{c}{} & \multicolumn{2}{c}{\textbf{Completion}} & \multicolumn{3}{c}{\textbf{Stability}} & \multicolumn{3}{c}{\textbf{Tracking Error}} \\
\cmidrule(r){1-1}\cmidrule(r){2-3} \cmidrule(r){4-6} \cmidrule(r){7-9}
Method & $\text{Succ}\tnote{1} \uparrow$ & $\text{Cont}\tnote{2} \downarrow$ &  $\text{Slip}\tnote{3} \downarrow$  & $\text{Air} \downarrow$ & $\text{Act}\tnote{4} \downarrow$ &  $E_\text{pos} \downarrow$ &  $E_{\text{vel}} \downarrow$  & $E_{\text{acc}} \downarrow$ \\

\cmidrule(r){1-1}\cmidrule(r){2-3} \cmidrule(r){4-6} \cmidrule(r){7-9}
\rowcolor{lightgray} 
\multicolumn{9}{l}{\textbf{(a) Ne Zha Pose }} \\
\cmidrule(r){1-1}\cmidrule(r){2-3} \cmidrule(r){4-6} \cmidrule(r){7-9}
H2O & 0 & 129.27 & 227.06 & 2.72 & 6.59 & 257.31 & 6.11 & 3.77 \\
OmniH2O & 0 & 146.19 & 219.04 & 5.03 & 4.60 & 102.38 & 4.70 & 3.41 \\  
\method & \textbf{97} & \textbf{0.02} & \textbf{72.76} & \textbf{0.69} & \textbf{0.46} & \textbf{74.13} & \textbf{2.94} & \textbf{1.65} \\

\cmidrule(r){1-1}\cmidrule(r){2-3} \cmidrule(r){4-6} \cmidrule(r){7-9}
\rowcolor{lightgray} 
\multicolumn{9}{l}{\textbf{(b) Single-leg Stand }} \\
\cmidrule(r){1-1}\cmidrule(r){2-3} \cmidrule(r){4-6} \cmidrule(r){7-9}
H2O & 0 & 172.71 & 236.28 & 3.05 & 8.76 & 478.23 & 7.23 & 4.25 \\
OmniH2O & 0 & 196.74 & 309.68 & 27.01 & 5.95 & 219.73 & 6.45 & 3.67 \\  
\method & \textbf{97} & \textbf{0.56} & \textbf{78.16} & \textbf{2.45} & \textbf{0.62} & \textbf{70.03} & \textbf{3.03} & \textbf{1.80} \\

\cmidrule(r){1-1}\cmidrule(r){2-3} \cmidrule(r){4-6} \cmidrule(r){7-9}
\rowcolor{lightgray} 
\multicolumn{9}{l}{\textbf{(c) Deep Squat }} \\
\cmidrule(r){1-1}\cmidrule(r){2-3} \cmidrule(r){4-6} \cmidrule(r){7-9}
H2O & \textbf{100} & \textbf{0.00} & 236.48 & 2.35 & 6.65 & 371.76 & 14.24 & 5.08 \\
OmniH2O & 99 & \textbf{0.00} & 141.20 & 0.94 & 1.46 & 101.40 & 7.04 & 2.84 \\  
\method & \textbf{100} & \textbf{0.00} & \textbf{77.93} & \textbf{0.12} & \textbf{0.77} & \textbf{62.28} & \textbf{5.58} & \textbf{2.31} \\

\bottomrule 
\end{tabular}

\begin{tablenotes}
Abbreviation for 
\item [1] \textit{Success Rate}
\item [2] \textit{Contact Mismatch}
\item [3] \textit{Slippage}
\item [4] \textit{Action Rate}
\end{tablenotes}
\vspace{2mm}
\end{threeparttable}
\endgroup}
\caption{
Simulation results of \method and baselines on additional 3 tasks.
}
\label{tab:additional_exp}
\vspace{-5mm}
\end{table}

To validate the necessity of our IMU-Centric Observation Perturbation component, we conduct simulation experiments on \texttt{Bruce Lee's Kick} comparing our coupled OU noise with other variants, including coupled Uniform noise as well as independent OU and independent Uniform (vanilla) noise. Here, \textit{coupled} means noise is first applied to the IMU's orientation observation and then the policy input observation is computed based on this noisy orientation, while \textit{independent} means noise is directly added to the policy input observation. The results in \Cref{tab:additional_imu_exp} show the coupled OU noise achieves the best overall performance, supporting its ability to better model real sensor noise and enhance policy robustness.

\begin{table}[H]
\centering
\begin{tabular}{lrrrrrrrr}
\toprule
\multicolumn{1}{c}{} & \multicolumn{2}{c}{\textbf{Completion}} & \multicolumn{3}{c}{\textbf{Stability}} & \multicolumn{3}{c}{\textbf{Tracking Error}} \\
\cmidrule(r){1-1}\cmidrule(r){2-3} \cmidrule(r){4-6} \cmidrule(r){7-9}
Method & $\text{Succ} \uparrow$ & $\text{Cont} \downarrow$ &  $\text{Slip} \downarrow$  & $\text{Air} \downarrow$ & $\text{Act} \downarrow$ &  $E_\text{pos} \downarrow$ &  $E_{\text{vel}} \downarrow$  & $E_{\text{acc}} \downarrow$ \\

\cmidrule(r){1-1}\cmidrule(r){2-3} \cmidrule(r){4-6} \cmidrule(r){7-9}

Independent Uniform & 98 & 0.02 & 162.63 & 6.10 & 4.40 & 105.21 & 4.81 & 3.12 \\
Independent OU & 98 & 0.03 & 140.68 & 5.34 & 3.35 & 99.22 & 4.55 & 2.82 \\ 
Coupled Uniform & \textbf{100} & \textbf{0.00} & 80.02 & 1.57 & 0.50 & 69.45 & 3.45 & \textbf{2.31} \\
Coupled OU (ours) & \textbf{100} & \textbf{0.00} & \textbf{76.44} & \textbf{1.28} & \textbf{0.46} & \textbf{67.18} & \textbf{3.31} & 2.33 \\ 
\bottomrule 
\end{tabular}
\vspace{2mm}
\caption{
Simulation results of \texttt{Bruce Lee's Kick} comparing different IMU noise variants.
}
\label{tab:additional_imu_exp}
\end{table}

\subsection{Failure Analysis of Baselines}
In our experiments, the failures of Baselines (H2O and OmniH2O) are primarily due to either falling or foot contact mismatch. The latter is treated as a failure, as contact with both feet indicates the single-leg task is not successfully completed. Refer to the \textit{Comparative Results} section on the project website for videos: the robot fails the \texttt{Swallow Balance} due to falling, and fails the \texttt{Ne Zha Pose} and \texttt{Single-Leg Stand} due to contact mismatch.
The low success rates of the baselines can be attributed to many factors, including—but not limited to—the absence of a contact penalty, overly small tracking tolerances, and inappropriate reward designs such as penalizing non-flat base orientations.

\end{document}